\documentclass{article}
\usepackage{ijcai23}

\usepackage{times}
\usepackage{soul}
\usepackage{url}
\usepackage[hidelinks]{hyperref}
\usepackage[utf8]{inputenc}
\usepackage[small]{caption}
\usepackage{graphicx}
\usepackage{amsmath}
\usepackage{amsthm}
\usepackage{booktabs}
\usepackage{algorithm}
\usepackage[switch]{lineno}

\usepackage{appendix}
\usepackage{bbm}
\usepackage{amsmath}
\usepackage{multirow}
\usepackage{textcomp}
\usepackage{tabularx}
\usepackage{dsfont}
\usepackage{booktabs}
\usepackage{enumerate}
\usepackage{nameref}
\usepackage{graphicx}
\usepackage{appendix}
\usepackage{subfig}
\usepackage{multirow, makecell}
\usepackage[T1]{fontenc}
\usepackage{algpseudocode}
\usepackage{graphicx}

\newcommand{\bag}{R}
\newcommand{\bagy}{Y}
\newcommand{\ins}{s}
\newcommand{\insy}{y}
\newcommand{\freq}{\rho}
\newcommand{\methodname}{Our method}

\usepackage[noabbrev]{cleveref}
\usepackage{textgreek}

    \makeatletter
\def\@fnsymbol#1{\ensuremath{\ifcase#1\or \dagger\or \ddagger\or
   \mathsection\or \mathparagraph\or \|\or **\or \dagger\dagger
   \or \ddagger\ddagger \else\@ctrerr\fi}}
    \makeatother

\title{A Noisy-Label-Learning Formulation for Immune Repertoire Classification and Disease-Associated Immune Receptor Sequence Identification}

\author{
Mingcai Chen $^1$$^*$\thanks{Work done during an internship at Tencent AI Lab. $\ddagger$ Corresponding author}\and
Yu Zhao $^{2}$$^*$\footnotemark[2]\and
Zhonghuang Wang $^{3,4}$\and 
Bing He $^{2}$\footnotemark[2]\And
Jianhua Yao $^2$\footnotemark[2]{}
\affiliations
$^1$State Key Laboratory for Novel Software Technology at Nanjing University, Nanjing University\\
$^2$Tencent AI Lab \\
$^3$National Genomics Data Center, Beijing Institute of Genomics, Chinese Academy of Sciences/China National Center for Bioinformation \\
$^4$College of Life Sciences, University of Chinese Academy of Sciences\\
\emails
chenmc@smail.nju.edu.cn,
louisyuzhao@tencent.com,
hebinghb@gmail.com, \\
wangzhonghuang17m@big.ac.cn,
jianhuayao@tencent.com
}

\begin{document}

\maketitle
\begin{abstract}
Immune repertoire classification, a typical multiple instance learning (MIL) problem, is a frontier research topic in computational biology that makes transformative contributions to new vaccines and immune therapies. However, the traditional instance-space MIL, directly assigning bag-level labels to instances, suffers from the massive amount of noisy labels and extremely low witness rate. In this work, we propose a noisy-label-learning formulation to solve the immune repertoire classification task. To remedy the inaccurate supervision of repertoire-level labels for a sequence-level classifier, we design a robust training strategy: The initial labels are smoothed to be asymmetric and are progressively corrected using the model's predictions throughout the training process. Furthermore, two models with the same architecture but different parameter initialization are co-trained simultaneously to remedy the known ``confirmation bias'' problem in the self-training-like schema. As a result, we obtain accurate sequence-level classification and, subsequently, repertoire-level classification. Experiments on the Cytomegalovirus (CMV) and Cancer datasets demonstrate our method's effectiveness and superior performance on sequence-level and repertoire-level tasks. Code available at \url{https://github.com/TencentAILabHealthcare/NLL-IRC}.

\end{abstract}
\section{Introduction}

Adaptive Immune Receptor Repertoires (AIRRs), including T-cell receptors (TCRs) and B-cell receptors (BCRs), can be mined for insights into the working mechanisms of the immune system, which is responsible for eliminating pathological microbes and toxic or allergenic proteins \cite{overview}.
There is a tremendous number of unique intra-individual and inter-individual adaptive immune receptor sequences to cover a broad space of potential antigens \cite{davis1988t}. 
For instance, the unique number of TCRs is estimated to maintain between $10^7$ and $10^8$ sampled from $10^{14}$ potential TCRs \cite{estimate}.
Upon antigen encounter, the B/T cells proliferate, producing daughter cells expressing the same antigen-specific AIRR sequences.
As a result of this clonal expansion, a fraction of the activated adaptive immune cells matures to a memory stage, leaving a signature of response \cite{human}.
Therefore, these signatures recognize antigens and record information on past and ongoing immune responses \cite{augmenting} that may serve as ultrasensitive biomarkers. 
Recently, studies about TCRs repertoire have succeeded in various medical applications, including predicting the presence of immunity after vaccination or infection, predicting the presence of disease, especially cancer and immune diseases, and designing antibody-based therapeutics \cite{therap}.
Moreover, many pioneer \textit{de novo} prediction methods have been proposed for immune repertoire classification \cite{DBLP:conf/nips/WidrichSPRGHBSG20,10.1371/journal.pone.0229569,sidhom2021deeptcr}, paving the way toward new vaccines and therapies.

\begin{figure*}[t]
    \centering
    \includegraphics[width=0.85\textwidth]{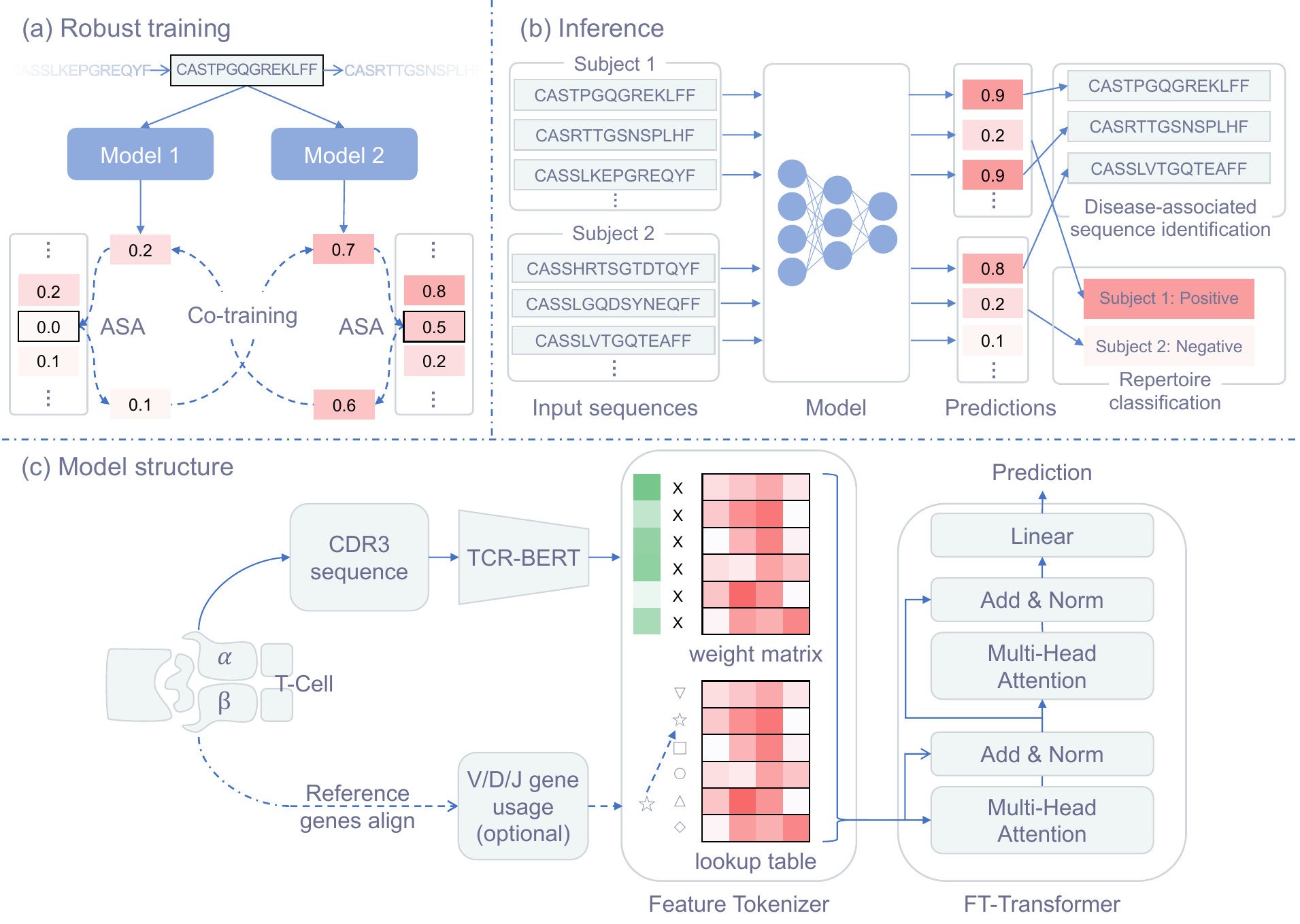} 
    \caption{(a) The robust training process with co-training and label correction. ASA stands for asymmetric self-adaptive label correction, which combines historical predictions and repertoire-level labels to supervise training. Dash lines indicate stop-gradient. (b). The inference process of our method, which can naturally perform disease-associated sequence identification and is ready for repertoire classification (c). Model architecture. Dash lines indicate "optional".} 
    \label{pipeline}
\end{figure*}


The immune repertoire classification can be formulated as a multiple instance learning problem.
An individual's repertoire is profiled as a bag of TCR sequences (instances), and the repertoire-level label indicates the immune status.
Current state-of-the-art studies solve this problem from the bag-classification perspective, i.e., generate a bag-level embedding for classification by aggregating the features of instances in a bag \cite{DBLP:conf/nips/WidrichSPRGHBSG20,10.1371/journal.pone.0229569}. 
We remark that such methodology suffers from two obstacles in practice:
\begin{itemize}
    \item As most TCR sequences are irrelevant to the disease of interest (i.e., the extremely low witness rate), the bag-classification methods, involving mostly irrelevant sequences when representing the repertoire, can hardly learn the immune responses of each individual sequence.
    \item The massive number of TCR sequences in each repertoire makes it computationally infeasible to feed all of them into a model. While sub-sampling the sequences inevitably filter out some of those specific to the antigens of interest, leading to sub-optimal performance.
\end{itemize}
Therefore, we suggest that the instance-space paradigm MIL is a more suitable solution.
However, the traditional instance-space MIL assigns the bag-level label to each instance, suffering from misleading supervision.

In this work, we propose to address the immune repertoire classification and associated sequence identification tasks using a noisy-label-learning formulation.
We develop a model for learning sequence-level (instance-level) concepts to directly characterize the biological relevance relationship between the target antigens and the TCR sequences.
The repertoire-level (bag-level) label is first utilized as a noisy substitution for the inaccessible ground-truth sequence-level label.
Only part of the obtained sequence-level labels is consistent with the ground truth because the repertoire status does not strictly correspond to the sequence-antigen association.
Therefore, the original MIL task is formulated as a noisy label problem \cite{algan2021image,song2020learning,DBLP:journals/corr/abs-2011-04406}, and we cope with it using robust training.

For the design of the robust training method, we first note that 
1). the healthy subjects would not have systemic proliferation of the disease-associated TCRs, and 
2). the patients would have more disease-specific TCRs, but a large proportion of the overall pool of TCRs is still irrelevant \cite{sidhom2021deeptcr} 
(we observe the low witness rate in Sec. \ref{sec_CMV_dataset} and empirically verify the domination of false positives in Sec. \ref{false_pos}).
To make the model learn from inaccurate labels, which mostly consist of false positives, an asymmetric self-adaptive pseudo-labeling strategy is proposed.
Modified from the label smoothing technique \cite{pereyra2017regularizing,lukasik2020does}, the initial training target is smoothed only when the sequence comes from a patient's repertoire.
The positive labels are progressively corrected using the predictions throughout the training process.
Having a self-training-like process, the model's error would accumulate over training.
To alleviate this, we employ co-training, i.e., maintaining two models and letting each learn from its peer.

After the robust training as shown in Fig. \ref{pipeline}(a), our model can produce sequences' level of relevance to the disease of interest.
The sequence-level classifier can also make inference on every TCR sequence in a repertoire, followed by a simple fusion strategy to form a repertoire-level prediction, as in Fig.~\ref{pipeline}(b).
In summary, our key contributions are:
\begin{itemize}
    \item We propose a noisy-label-learning formulation that utilizes repertoire-level concepts to train a sequence-level classifier for both immune repertoire classification and disease-associated TCR sequence identification.
    \item We design a robust training algorithm for the sequence classifier. Noisy labels are refined using the asymmetric self-adaptive correction strategy. What's more, to alleviate the confirmation bias in the self-training, we deploy co-training to produce better training signals.
    \item We verify our method on the CMV and Cancer dataset.
    The results demonstrate our method's superiority in immune repertoire classification and disease-associated sequence identification. 
\end{itemize}

\section{Method}

\subsection{Problem Setting}
Assuming $N$ Individuals' repertoires are denoted as $\{\bag_1,\ldots,\bag_N\}$ and the corresponding repertoire-level concepts are denoted as $\{\bagy_1,\ldots,\bagy_N\}$.
$\bagy \in \{0,1\}$, where 1 represents that subject is positive, and 0 represents negative 
(To simplify the discussion, we assume the task is binary classification, although our method can be easily extended to multi-class or regression).
Each repertoire consists of a large amount of sequences $\{\ins_1,\ldots,\ins_M\}$ and their corresponding frequency $\{\freq_1,\ldots,\freq_M\}$.
Each sequence also has underlying corresponding sequence-level labels $\insy$, where $\insy \in \{0,1\}$, where 1 represents that the TCR is associated with the disease of interest, and 0 represents that it's not.
We assume the sequence-level label is unavailable during training.
Only a relatively small amount of them are known and used for evaluation.
We have two objectives:
\begin{itemize}
    \item Repertoire classification, i.e., a mapping function $F: \bag \mapsto \bagy$.
    \item Disease-associated sequence identification, i.e., a mapping function $f: \ins \mapsto \insy$
\end{itemize}

\subsection{Our Method}
Many previous methods model the repertoire classification problem as MIL, and some of them identify disease-associated sequence as a side feature.
Our work attempts to learn the antigen-driven responses directly.
However, the corresponding antigen of most of the sequences within repertoires is unknown.
It's hard to obtain enough labeled sequences for the training of a deep model.
Considering the characteristics of this problem, we transform the repertoire-level labels to noisy sequence-level labels, i.e., letting $\bagy = \tilde{\insy}$.
$\tilde{\insy}$ is the noisy sequence label and may not be the same as the ground-truth $\insy$, especially when the sequence is from the repertoire of a patient. 
We are to train a sequence-level model $f: s \mapsto \insy$, which predicts whether an input sequence belongs to a sequence-level concept, i.e., disease-irrelevant or disease-associated.


\subsubsection{Model Architecture}
We use a simple adaption of the Transformer architecture \cite{DBLP:conf/nips/VaswaniSPUJGKP17} named FT-Transformer \cite{DBLP:conf/nips/GorishniyRKB21}.
It is designed for tabular data and can take both categorical and numerical data as input.
To better extract meaningful information from the complex CDR3 sequences, we use TCR-BERT \cite{DBLP:conf/naacl/DevlinCLT19} -- a BERT-like pre-trained model for TCR sequences -- to obtain feature representation. 
Then the categorical V/D/J genes and feature representations of CDR3 sequences are fed into the FT-transformer.
The transformer-based model has been extensively used, so we do not elaborate on the details here.
Please refer to \cite{DBLP:conf/nips/VaswaniSPUJGKP17,DBLP:conf/nips/GorishniyRKB21}.
The structure is shown in Fig. \ref{pipeline} (c).

\subsubsection{Training Schema}
After using the repertoire-level label $Y$ as noisy label $\tilde{\insy}$ for training, conventional Empirical Risk Minimization (ERM) is deemed to fail to obtain a generalizable model.
Therefore, we design a training schema that is robust to noisy labels.

As found in several previous studies \cite{DBLP:journals/corr/RolnickVBS17,DBLP:conf/aistats/LiSO20,DBLP:conf/aaai/GuanGDH18}, model predictions magnify useful information in data.
The self-adaptive training \cite{huang2020self} thus progressively corrects the training targets using the Exponential Moving Average (EMA):
\begin{equation}
\label{SA}
    t_{i,j}= \alpha \times t_{i,j-1} +(1-\alpha)\times \mathrm{p}_{}(y_i=1 \mid x_i,\theta_{j-1})
\end{equation}
where $t_{i,j}$ is the training target for $i$-th sample in $j$-th epoch.
$\alpha$ is a hyper-parameter that controls how quickly the history predictions replace the training targets.

Label smoothing is a commonly used technique in robust training. 
In a binary classification task, it replaces the one-hot encoded label vector with a mixture of the original one and the uniform distribution, as shown in Eq. \ref{ls}.
$\beta$ is a hyper-parameter that controls how much the training targets are smoothed.
Considering the domination of false positive samples in our problem, we modify the original form to only smooth the target label when the sample is positive.
It is used to initialize the training target as Eq. \ref{als} and integrated with the self-adaptive strategy.

\begin{equation}
\label{ls}
    t_i=\left\{
        \begin{array}{lc}
        1-\beta, & y_i=1 \\
        \beta,   & y_i= 0\\
    \end{array}
    \right.
\end{equation} 
\begin{equation}
\label{als}
    t_{i,0}=\left\{
        \begin{array}{lc}
        1-\beta, & y_i=1 \\
        0,   & y_i= 0\\
    \end{array}
    \right.
\end{equation}

We use the soft cross-entropy as the only training loss after correcting the training target,
\begin{equation}
\begin{aligned}
    \mathcal{L}=-\sum_i t_{i,j} \mathrm{log}(\mathrm{p}_{}(y_i=1 \mid x_i,\theta_{j}))\\+(1-t_{i,j}) \mathrm{log}(\mathrm{p}_{}(y_i=0 \mid x_i,\theta_{j}))
\end{aligned}
\end{equation}
The model is trained for several epochs before using the corrected training targets (the number of warm-up epochs is a hyper-parameter and tuned with others).

After introducing pseudo-labels, the known ``confirmation bias'' problem \cite{tarvainen2017mean,arazo2020pseudo} appears: the model's wrong predictions would be used to guide subsequent training, making the error accumulate.
To address this problem, we introduce co-training.
Specifically, two models with the same architecture but different random parameter initialization are trained simultaneously.
Instead of self-training, The training targets generated by one model, as in Eq. \ref{SA}, are to train its peer.
The final results are their ensemble (the average of their outputs probabilities).

\subsubsection{Repertoire Classifier}
The repertoire classifier $F$ is constructed using the sequence classifier $f$.
One repertoire consists of multiple triplets: CDR3 sequence, V/D/J gene, and productive frequency.
The CDR3 sequence and V/D/J gene are the input of the sequence-level model, which produces a prediction value in [0,1].
The score the repertoire-level model assigns to a repertoire $R$ is calculated as
\begin{equation}
    F(R)=\sum_{(\ins,\freq)\in R} \freq f(\ins )
\end{equation}
where $\freq$ is the frequency of the corresponding sequence.
In this way, the bigger the probability the sequences are associated with the disease, the more likely the subject is exposed to specific antigens.
The more frequent a sequence appears in a repertoire, the more it would affect the repertoire-level classification.

\section{Experiments}
\subsection{Data Collection}
A T-cell receptor is profiled by the CDR3 sequence, V/D/J gene usage, and its productive frequency.
The frequency is the read count of the productive CDR3 Amino Acid (AA) sequence over the total number of productive reads in each repertoire.
We then introduce both datasets.


\subsubsection{Cytomegalovirus} 
\label{sec_CMV_dataset}
We use the CMV dataset \cite{emerson2017immunosequencing}, which consists of 785 individuals' repertoires and their corresponding serostatus.
We include a total of 685 repertoires with known CMV serostatus and sequence abundance, which are evenly and randomly split into 5 folds for cross-validation. 
In each run, 3 out of 5 folds are used for training, 1 for validation, and 1 for test.
Because the whole repertoire is too big, we sub-sample 0.1\% most frequently appeared sequences in the repertoire.
It is worth noting that, our method can take all sequences for training, while others are limited by the memory cost.
We sub-sample repertoires in exchange for efficiency.

\begin{figure}
    \centering
    \includegraphics[width=0.9\columnwidth]{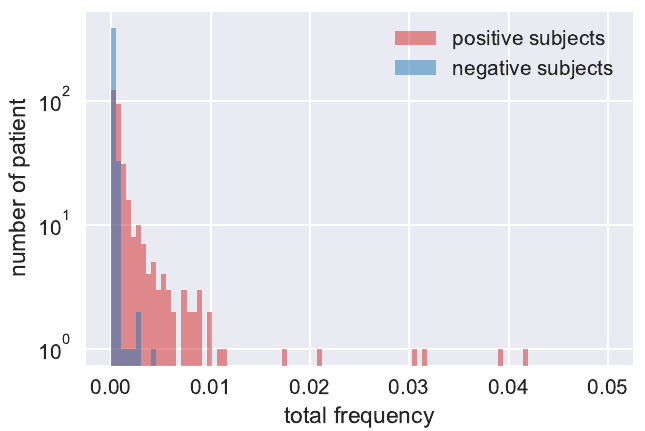} 
    \caption{The sum of occurrence frequency of the discovered CMV-associated sequences appearing in positive and negative subjects.} 
    \label{hits}
\end{figure}

Moreover, we use a total of 1,085 CMV-associated TCRs identified by many previous studies, including \cite{TCR1,TCR2}, through \textit{in vitro} methods and curated in \cite{emerson2017immunosequencing} as the ground-truth positive labels for the disease-associated sequence identification task. 
Note that we only use these labels for evaluation and don't access them during training.
We draw the occurrence frequency of the discovered CMV-associated sequences in positive and negative patients in Fig. \ref{hits}, which verifies: 1.) Negative repertoires' also have CMV-associated sequences, and 2). Positive repertoires tend to have more CMV-associated sequences, and 3.) The witness rate is low (even the positive repertoires have less than 5\% sequences that match the known CMV-associated sequences in their repertoire).
It is also worth noting that this set does not cover all CMV-associated sequences in patients' repertoires.
Some positive repertoires contain sequences not in the set.
On top of the 5 folds cross-validation schema in the repertoire classification task, the intersected sequences between the test fold and the 1,085 known CMV-associated TCRs are used as positive test samples, and other sequences in the healthy donors are used as negative samples.

\subsubsection{Cancer} 
We use the dataset curated in \cite{beshnova2020novo}. 
Its training set consists of 40,000 cancer-associated TCR\textbeta~CDR3 sequences along with 30,000 for controls.
Its test set includes 10,000 cancer-associated TCR\textbeta~CDR3 sequences and 10,000 for controls.
The cancer-associated sequences are collected by excluding T cell receptors in healthy donors from that in patients.
However, there could be many false positives in the labeling process, which still presents us with a label noise problem. 
Following \cite{beshnova2020novo}, we randomly choose 1/3 of the training set for validation and report the accuracy on the test set.

\begin{figure*}
    \centering
    \includegraphics[width=0.95\textwidth]{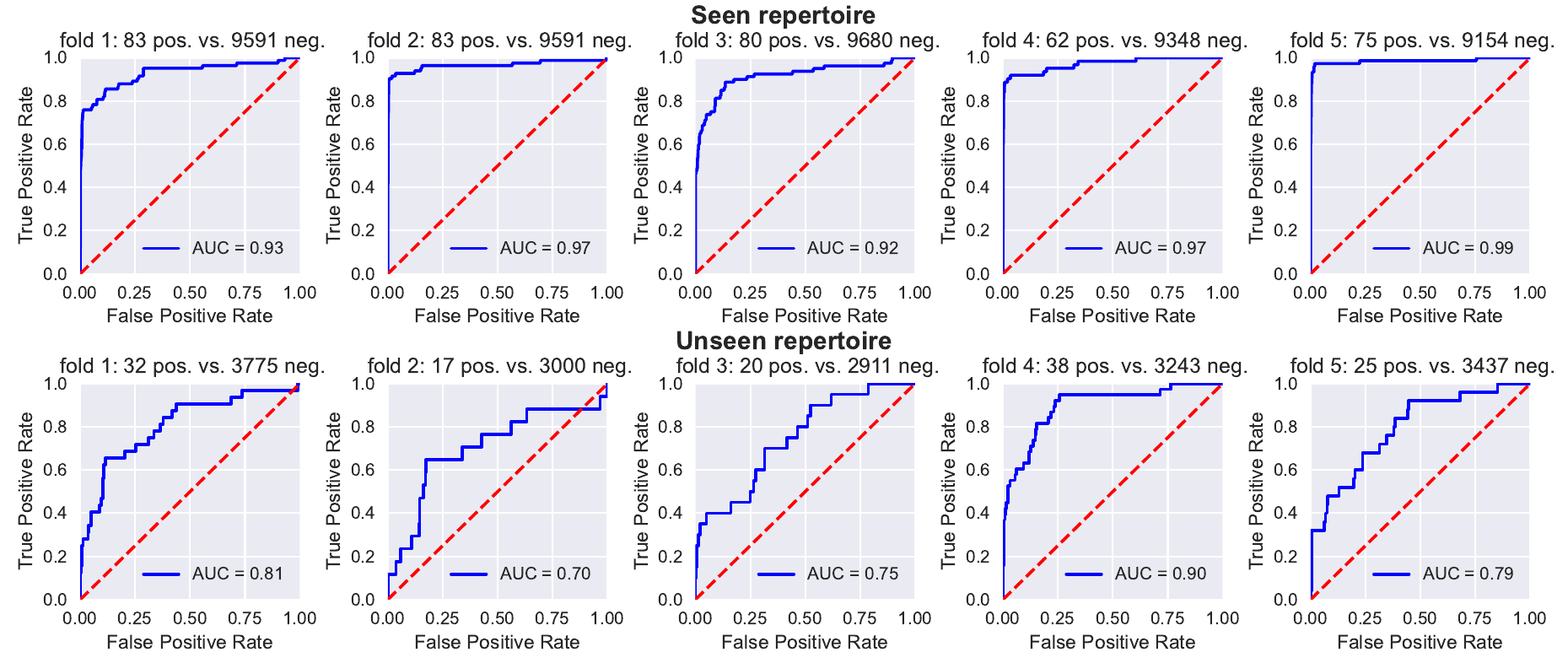} 
    \caption{The ROC of associated-sequence identification. Only trained by repertoire-level labels, our model identifies the most associated sequences. In the second row, we let the model make predictions on repertoires that are outside of the training dataset.} 
    \label{associated_seqs}
\end{figure*}

\subsection{Hyper-Parameters} 
For the experiments on the CMV dataset, We choose the hyper-parameters using the validation set. Specifically, $\alpha$ in EMA is set to 0.99, and $\beta$ in label smoothing is set to 0.7. 
In terms of the model structure, there is 1 layer of 1-head attention, the dimension of the embedding token is 16, and the dropout rate is 0.
The optimizer is Adam optimizer \cite{DBLP:journals/corr/KingmaB14} with a learning rate of 0.005 and a batch size of 256.
We train the model for 15 epochs using the original targets to warm the model up.
The validation set is used for early stopping.
For the experiments on the Cancer dataset, We choose the hyper-parameters using the validation set. Specifically, $\alpha$ in EMA is set to 0.95, and $\beta$ in label smoothing is set to 0.4. 
In terms of the model structure, there are 2 layers of 4-head attention, the dimension of the embedding token is 192, and the dropout rate is 0.1.
The optimizer is Adam optimizer \cite{DBLP:journals/corr/KingmaB14} with a learning rate of 0.0005 and a batch size of 256.
We train the model for 8 epochs using the original targets to warm the model up.
The validation set is used for early stopping.

\subsection{Experiments on CMV Dataset}
We first verify the repertoire classification and disease-associated TCR sequence abilities of our model on the CMV dataset.
For the classification task, we compare our method with:
\begin{itemize}
    \item Logistic regression as a baseline linear classifier.
    \item Burden test in \cite{emerson2017immunosequencing} as a baseline statistical method.
    \item Logistic multiple instance learning (Log. MIL)  \cite{10.1371/journal.pone.0229569} applies logistic regression to each k-mer representation.
    \item DeepRC \cite{DBLP:conf/nips/WidrichSPRGHBSG20} is a modern Hopfield network that models the repertoire classification as MIL.
\end{itemize}
It can be seen from Table \ref{cmvauc} that even though our method is only fed with sequence and has no access to the whole repertoire, competitive results are achieved.

We then evaluate the sequence-level predictions to explore whether our model learns the pair-wise TCR–disease relationship.
As shown in the first row in Fig. \ref{associated_seqs}, our model surprisingly assigns accurate association scores on the previously known CMV-associated sequences that are identified by \cite{emerson2017immunosequencing}.
It demonstrates prediction accuracy competitive with experimental results in a majority of cases.
It is also worth noting that there are a small number of false negatives in the test set because healthy donors may still have disease-associated sequences.
Therefore, the reported AUC values tend to be lower than the real performance. 
We also release those sequences with high confidence but not in the associated sequences set in the supplementary material for future experimental verification. 

We further test our model on a more challenging task, i.e., on repertoires that are not even seen during training. The results in the second row of Fig. \ref{associated_seqs} show that it is also possible to make predictions on unseen hypervariable TCR sequences.

\begin{table}
     \centering
     \resizebox{\columnwidth}{!}{ %
     \begin{tabular}{l|rrr}
     \toprule
        Method                  & AUC $\pm$ std           &  F1 score $\pm$ std      & Acc. $\pm$ std   \\ \midrule
        Logistic regression     & 60.70 $\pm$ 5.8         &  24.4 $\pm$ 20.6  & 59.0 $\pm$ 1.9\\
        Burden test             & 69.90 $\pm$ 4.1         &   -            &     -          \\
        Log. MIL (KMER)         & 58.20 $\pm$ 6.5         &  11.8 $\pm$ 26.4 &   51.5 $\pm$ 5.8 \\
        Log. MIL (TCR\textbeta) & 51.50 $\pm$ 7.3         &  0.0 $\pm$ 0.0  &   54.1 $\pm$ 3.9\\
        DeepRC                  & 83.10 $\pm$ 2.2&  72.6 $\pm$ 5.0   &  72.7 $\pm$ 4.9\\ 
        \methodname             & \textbf{83.24} $\pm$ 2.8&  \textbf{75.9} $\pm$ 2.4   &  \textbf{76.1} $\pm$ 2.4\\ \bottomrule
     \end{tabular}
     }
     \caption{Performance on CMV dataset. Other methods' results are from DeepRC. Acc. stands for Accuracy. Best results are in bold. Note that for the calculation of the F1 score and accuracy, we select a threshold value based on the validation set.}
     \label{cmvauc}
\end{table}

\begin{figure*}
    \centering
    \includegraphics[width=0.85\textwidth]{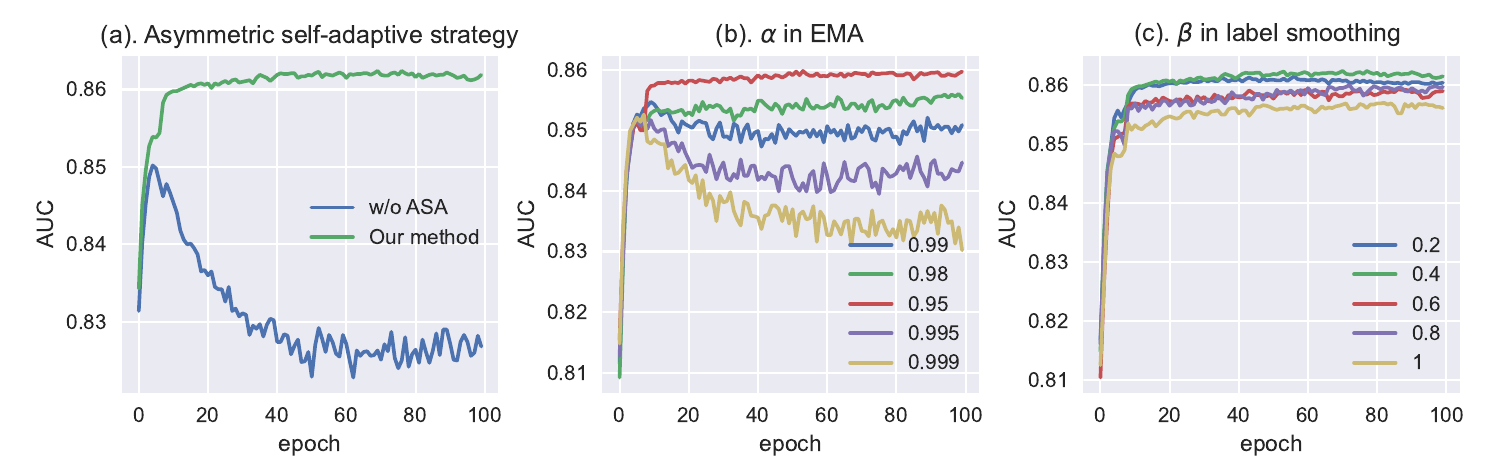} 
    \caption{\methodname's performance (on the validation set) varies with the (a). label correction technique, (b). $\alpha$ in EMA, and (c). $\beta$ in label smoothing.} 
    \label{ablation}
\end{figure*}

\begin{figure}
    \centering
    \includegraphics[width=\columnwidth]{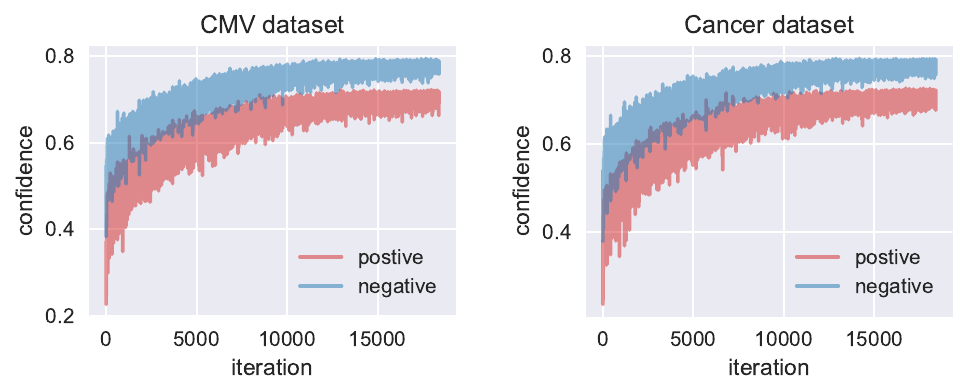} 
    \caption{The average confidence on two classes' samples. } 
    \label{conf}
\end{figure}

\subsection{Experiments on Cancer Dataset}

We further verify the disease-associated TCR sequence identification ability of our model on the Cancer dataset.
We mainly compare our method on the cancer dataset with DeepCAT, which deploys CNN models after PCA encoding of the AA biochemical indices, AutoCAT \cite{DBLP:journals/bioinformatics/WongL22}, which extends DeepCAT by allowing the use of immune repertoire samples, and TCR-BERT + Linear Regression, i.e., the linear classifier on top of the embedding from TCR-BERT.

DeepCAT also attempts to directly model the sequence-disease relationship. 
However, we suggest that the constructed Cancer dataset is noisy by making the observation in Fig. \ref{conf}.
DeepCAT trains one model for each length, and our method takes all sequences as input.
Thus, besides the overall AUC, we also report the prediction results of our method on 12-17 AA long test sequences for comparison with DeepCAT. 

From Table \ref{deepcatauc}, our method consistently outperforms DeepCAT across all sequence lengths.
The effectiveness of our robustness training method verifies the existence of noisy labels in the Cancer dataset.

\begin{table}
     \centering
     \resizebox{\columnwidth}{!}{ %
     \begin{tabular}{l|cccccc}
         \toprule
 		 Seq. length & 12 & 13 & 14 & 15 & 16 & Avg.\\ \midrule
 		 DeepCAT & 59 & 53 & 62 & 76 & 86 & 80\\
 		 AutoCAT & 73 & 69 & 70 & 70 & 71 & 71 \\
              TCR-BERT + LR & 83.21 & 83.37 & 82.06 & 82.91 & 82.06 & 84.28 \\
 		 \methodname & \textbf{89.36} & \textbf{88.53} & \textbf{87.18} & \textbf{87.99} & \textbf{87.16}  & \textbf{89.40}\\ \bottomrule
     \end{tabular}
     }
     \caption{AUC (\%) on Cancer dataset. DeepCAT's results are from its original paper. 
     Note the last column is the AUC of all sequences, which includes other lengths. Best results are in bold.}
     \label{deepcatauc}
\end{table}

\subsection{Explore the Distribution of Noisy Labels}
\label{false_pos}
To explore whether our assumption about the noisy labels is mostly from the positive class, we try to empirically prove it in this section.

As shown in \cite{arpit2017closer}, deep models would learn easy and simple patterns first before fitting the noise.
The conclusion is reached by training a deep model on a noisy dataset and making the observation that the clean samples are fitted before the noisy samples.
Similarly, we train the model only using ERM and original labels and observe the model's prediction confidence (the softmax probabilities on the corresponding class) on samples in negative and positive classes, respectively.
As shown in Fig. \ref{conf}, the model starts to produce higher confidence on negative samples from the early training stage.
Besides, the training iterations for the model to overfit the positive samples are more than that of the negative samples.
Note that the original labels are balanced on both datasets.
So this observation verifies that there are more noisy positive labels than noisy negative labels.

\subsection{Ablation Study}
To verify the effectiveness of different components of our method, we conduct ablation experiments on the CMV and Cancer datasets. We design two variants of our method:
\begin{itemize}
    \item w/o ASA: removing the asymmetric self-adaptive training strategy.
    \item w/o CT: removing the co-training strategy.
\end{itemize}

It can be seen from Table \ref{table_ab} that the two components both bring performance gain.
It is worth noting that the asymmetric self-adaptive strategy brings nearly no extra computation, facilitating the adaptation of our method.
What's more, the asymmetric self-adaptive technique ensures more stable training, as shown in the training curve in Fig. \ref{ablation}(a).

\begin{table}
     \centering
     \begin{tabular}{l|cc}
         \toprule
 		 Ablation    & CMV &  DeepCAT  \\ \midrule
 		 \methodname & 83.24 $\pm$ 0.03 & 86.23 $\pm$ 0.07 \\
 		 w/o ASA     & 82.50 $\pm$ 0.03 & 85.01 $\pm$ 0.04 \\
 		 w/o CT      &  82.84 $\pm$ 0.03 & 85.07 $\pm$ 0.08 \\ \bottomrule
     \end{tabular}
     \caption{Ablation study. Results (AUC (\%) $\pm$ std) on CMV and DeepCAT are reported. Best results are in bold.}
    \label{table_ab}
\end{table}

We also compare the choice of two important hyper-parameters in Fig. \ref{ablation}(b) and (c).
It can be found that the $\alpha$ in EMA should be carefully chosen.  
Increasing $\alpha$ would make the original labels remain longer, gradually degenerating into the standard ERM.
While decreasing $\alpha$ would make the training targets quickly approximate the model's prediction and the training loss close to zero.

\section{Related Work}
\subsection{Repertoire Classification and Disease-Associated Sequences Detection}
Repertoire-based machine learning methods and applications focus on the classification and prediction of a donor's immune status, such as disease status, recent vaccination, or past exposure to a particular pathogen. 
Thus, the main application area is immunodiagnostics. 

In order to capture antigen-associated sequences at a repertoire level, there exists a range of approaches:
1). Public adaptive immune receptor-based classification. Emerson et al. \cite{emerson2017immunosequencing} identified TCRβ sequences associated with cytomegalovirus (CMV) positive serostatus using Fisher's exact test. 
Only 164 sequences were found in positive individuals, demonstrating that the detectable signal of pathogen exposure may be small. 
A similarly small number of TCRβ sequences that are associated with systemic lupus erythematosus and rheumatoid arthritis are distinguished from healthy individuals \cite{lupus}.
2). Sequence similarity cluster-based classification. 
Using GIANA \cite{GIANA} to cluster large-scale TCR datasets provided candidate antigen-specific receptors and provided a new solution to repertoire classification. 
This study is significant in that it identified sequence clusters that distinguish individuals.
3). K-mer-based classification. Many repertoire-based approaches rely on k-mer encoding, where each repertoire represents a 100-element vector corresponding to the distribution of CDR3β k-mers in the repertoire \cite{tracking}. 
They first cluster them in accordance with their physicochemical properties before classifying individuals using SVM approaches. 
4). MIL–based classification methods predict sequences associated with the antigen within a repertoire and then compute a repertoire-level label from the set of sequence-level predictions. 
Ostmeyer et al. \cite{stat} first developed an approach that utilized MIL for diagnosing multiple sclerosis (MS), which is traditionally difficult to distinguish from other neurological diseases. 
In addition, they have demonstrated that ovarian cancer-associated repertoires from women can be distinguished from cancer-free women \cite{biophysi}. 
With the aim of advancing cancer screening, Beshnova et al. \cite{beshnova2020novo} extended this work to identify cancer-associated TCRs. 
Recently, Widrich et al. \cite{DBLP:conf/nips/WidrichSPRGHBSG20} developed an attention-based MIL deep learning approach that accurately predicts CMV serostatus and additionally used a transformer-like attention mechanism extracting those sequence motifs that are connected to disease classes.

\subsection{Multiple Instance Learning}
Multiple instance learning is a typical weakly supervised learning problem. Current MIL algorithms have three paradigms, i.e., instance-space paradigm, bag-space paradigm, and embedded-space paradigm \cite{wang2018revisiting}, categorized based on which level to learn the discriminative information (instance-level or bag-level).
The instance-space paradigm focus on learning instance information, which learns the instance classifier at the first stage and then achieves the bag-level classifier by aggregating instance-level results. The bag-space paradigm and embedded-space paradigm (entire-bag-embedding MIL), on the other hand, extract discriminative information from the whole bag. In immune repertoire classification, current existing methods mostly belong to the embedded-space paradigm \cite{DBLP:conf/nips/WidrichSPRGHBSG20,10.1371/journal.pone.0229569}. 
However, the nature of the huge instance size ($10^7-10^8$) and extremely low witness rate of the immune repertoire challenge the performance of entire-bag-embedding methods in practice,

\subsection{Learning with Noisy Labels}
Most deep learning methods assume perfect label information, which is impractical in many real-world scenarios.
Learning with Noisy Label (LNL) aims to alleviate the problem since weak supervision can be obtained more efficiently. 

Recently, several LNL methods have attracted much attention and obtained much better empirical performance \cite{song2020learning}. 
Sample selection methods select possible mislabeled samples and exclude them from training.
For example, \cite{jiang2018mentornet} uses a pre-trained network to filter samples with big losses, which tend to be noisy, and a student only trained by the selected clean dataset.
\cite{han2018co} maintains two networks, and each selects certain small-loss samples and uses them as training signals for its peer.
On top of sample selection, one can also utilize possible mislabeled data with semi-supervised learning techniques.
For example,  DivideMix \cite{li2020dividemix} first divides samples into clean and noisy using the Gaussian mixture model.
Then it deploys a semi-supervised learning method to leverage both subsets.
RoCL \cite{zhou2021robust} uses the EMA of loss value for sample selection and label correction. 
Furthermore, it uses supervised training for selected clean samples and uses semi-supervised learning for selected noisy samples.

\section{Broader Impact}
\methodname~is a deep learning method and is not expected to be applied to shifted distribution data or out-of-distribution data.
We leave this as a starting point for future work.
More importantly, our method is not meant to take part in patients' treatment or therapy directly.
Therefore, no human beings are at risk of being treated without reflecting.
Our method's applications include medical or biological experiments or help to preliminarily investigate the immune repertoire data, gain insights, and possibly derive treatments.

\section{Conclusion}
We rethink the modeling of the immune repertoire classification problem in this paper. 
After suggesting several defects of the conventional MIL, the new formulation -- a straightforward sequence classifier accompanied by robust training -- is proposed for 
 both more reasonable modeling and better prediction performance.
For the inaccurate sequence-level labels transformed from repertoire-level labels, a label correction method mainly targeted at the false positives is proposed.
As a result, our model achieves superior repertoire classification results.
What's more, it successfully identifies most disease-associated sequences from previous biology studies.
We envision the work could encourage future research on how to gain insights into the TCR–disease relationship without accurate supervision.

\section*{Contribution Statement}
Mingcai Chen and Yu Zhao equally contributed to this work. The detailed contribution of all authors are summarized as follows: Conceptualization: Y. Z., B. H. and J. Y.; methodology: M. C. and Y. Z.; investigation and analysis: M. C., Y. Z., B. H., Z. W. and J. Y.; writing: M. C., Y. Z., B.H., Z. W. and J. Y.; supervision: Y. Z., B. H. and J. Y.

\bibliographystyle{named}
\bibliography{sample}

\begin{thebibliography}{}

\bibitem[\protect\citeauthoryear{Algan and Ulusoy}{2021}]{algan2021image}
G{\"o}rkem Algan and Ilkay Ulusoy.
\newblock Image classification with deep learning in the presence of noisy
  labels: A survey.
\newblock {\em Knowledge-Based Systems}, 215:106771, 2021.

\bibitem[\protect\citeauthoryear{Arakaki \bgroup \em et al.\egroup
  }{2010}]{TCR1}
Atsushi Arakaki, Kaori Ooya, Yasuto Akiyama, Masahito Hosokawa, Masaru
  Komiyama, Akira Iizuka, Ken Yamaguchi, and Tadashi Matsunaga.
\newblock Tcr-β repertoire analysis of antigen-specific single t cells using a
  high-density microcavity array.
\newblock {\em Biotechnology and bioengineering}, 106:311--8, 06 2010.

\bibitem[\protect\citeauthoryear{Arazo \bgroup \em et al.\egroup
  }{2020}]{arazo2020pseudo}
Eric Arazo, Diego Ortego, Paul Albert, Noel~E O’Connor, and Kevin McGuinness.
\newblock Pseudo-labeling and confirmation bias in deep semi-supervised
  learning.
\newblock In {\em 2020 International Joint Conference on Neural Networks
  (IJCNN)}, pages 1--8. IEEE, 2020.

\bibitem[\protect\citeauthoryear{Arpit \bgroup \em et al.\egroup
  }{2017}]{arpit2017closer}
Devansh Arpit, Stanis{\l}aw Jastrz{\k{e}}bski, Nicolas Ballas, David Krueger,
  Emmanuel Bengio, Maxinder~S Kanwal, Tegan Maharaj, Asja Fischer, Aaron
  Courville, Yoshua Bengio, et~al.
\newblock A closer look at memorization in deep networks.
\newblock In {\em International Conference on Machine Learning}, pages
  233--242. PMLR, 2017.

\bibitem[\protect\citeauthoryear{Arstila \bgroup \em et al.\egroup
  }{1999}]{estimate}
T~Arstila, A~Casrouge, Vr~Baron, J~Even, Jean Kanellopoulos, and Philippe
  Kourilsky.
\newblock A direct estimate of the human alphabeta t cell receptor diversity.
\newblock {\em Science (New York, N.Y.)}, 286:958--61, 11 1999.

\bibitem[\protect\citeauthoryear{Babel \bgroup \em et al.\egroup }{2008}]{TCR2}
Nina Babel, Gordon Brestrich, Lukasz Gondek, Arne Sattler, Marcin Wlodarski,
  Nina Poliak, Nicole Bethke, Andreas Thiel, Markus Hammer, Petra Reinke, and
  Jaroslaw Maciejewski.
\newblock Clonotype analysis of cytomegalovirus-specific cytotoxic t
  lymphocytes.
\newblock {\em Journal of the American Society of Nephrology : JASN},
  20:344--52, 10 2008.

\bibitem[\protect\citeauthoryear{Beshnova \bgroup \em et al.\egroup
  }{2020}]{beshnova2020novo}
Daria Beshnova, Jianfeng Ye, Oreoluwa Onabolu, Benjamin Moon, Wenxin Zheng,
  Yang-Xin Fu, James Brugarolas, Jayanthi Lea, and Bo~Li.
\newblock De novo prediction of cancer-associated t cell receptors for
  noninvasive cancer detection.
\newblock {\em Science translational medicine}, 12(557):eaaz3738, 2020.

\bibitem[\protect\citeauthoryear{Brown \bgroup \em et al.\egroup
  }{2019}]{augmenting}
Alex Brown, Igor Snapkov, Rahmad Akbar, Milena Pavlović, Enkelejda Miho, Geir
  Sandve, and Victor Greiff.
\newblock Augmenting adaptive immunity: Progress and challenges in the
  quantitative engineering and analysis of adaptive immune receptor
  repertoires.
\newblock {\em Molecular Systems Design \& Engineering}, 4, 07 2019.

\bibitem[\protect\citeauthoryear{Davis and Bjorkman}{1988}]{davis1988t}
Mark~M Davis and Pamela~J Bjorkman.
\newblock T-cell antigen receptor genes and t-cell recognition.
\newblock {\em Nature}, 334(6181):395--402, 1988.

\bibitem[\protect\citeauthoryear{Devlin \bgroup \em et al.\egroup
  }{2019}]{DBLP:conf/naacl/DevlinCLT19}
Jacob Devlin, Ming{-}Wei Chang, Kenton Lee, and Kristina Toutanova.
\newblock {BERT:} pre-training of deep bidirectional transformers for language
  understanding.
\newblock In Jill Burstein, Christy Doran, and Thamar Solorio, editors, {\em
  Proceedings of the 2019 Conference of the North American Chapter of the
  Association for Computational Linguistics: Human Language Technologies,
  {NAACL-HLT} 2019, Minneapolis, MN, USA, June 2-7, 2019, Volume 1 (Long and
  Short Papers)}, pages 4171--4186. Association for Computational Linguistics,
  2019.

\bibitem[\protect\citeauthoryear{Emerson \bgroup \em et al.\egroup
  }{2017}]{emerson2017immunosequencing}
Ryan~O Emerson, William~S DeWitt, Marissa Vignali, Jenna Gravley, Joyce~K Hu,
  Edward~J Osborne, Cindy Desmarais, Mark Klinger, Christopher~S Carlson,
  John~A Hansen, et~al.
\newblock Immunosequencing identifies signatures of cytomegalovirus exposure
  history and hla-mediated effects on the t cell repertoire.
\newblock {\em Nature genetics}, 49(5):659--665, 2017.

\bibitem[\protect\citeauthoryear{Farber \bgroup \em et al.\egroup
  }{2013}]{human}
Donna Farber, Naomi Yudanin, and Nicholas Restifo.
\newblock Human memory t cells: Generation, compartmentalization and
  homeostasis.
\newblock {\em Nature reviews. Immunology}, 14, 12 2013.

\bibitem[\protect\citeauthoryear{Gorishniy \bgroup \em et al.\egroup
  }{2021}]{DBLP:conf/nips/GorishniyRKB21}
Yury Gorishniy, Ivan Rubachev, Valentin Khrulkov, and Artem Babenko.
\newblock Revisiting deep learning models for tabular data.
\newblock In Marc'Aurelio Ranzato, Alina Beygelzimer, Yann~N. Dauphin, Percy
  Liang, and Jennifer~Wortman Vaughan, editors, {\em Advances in Neural
  Information Processing Systems 34: Annual Conference on Neural Information
  Processing Systems 2021, NeurIPS 2021, December 6-14, 2021, virtual}, pages
  18932--18943, 2021.

\bibitem[\protect\citeauthoryear{Greiff \bgroup \em et al.\egroup
  }{2020}]{therap}
Victor Greiff, Gur Yaari, and Lindsay Cowell.
\newblock Mining adaptive immune receptor repertoires for biological and
  clinical information using machine learning.
\newblock {\em Current Opinion in Systems Biology}, 24:109--119, 12 2020.

\bibitem[\protect\citeauthoryear{Guan \bgroup \em et al.\egroup
  }{2018}]{DBLP:conf/aaai/GuanGDH18}
Melody~Y. Guan, Varun Gulshan, Andrew~M. Dai, and Geoffrey~E. Hinton.
\newblock Who said what: Modeling individual labelers improves classification.
\newblock In Sheila~A. McIlraith and Kilian~Q. Weinberger, editors, {\em
  Proceedings of the Thirty-Second {AAAI} Conference on Artificial
  Intelligence, (AAAI-18), the 30th innovative Applications of Artificial
  Intelligence (IAAI-18), and the 8th {AAAI} Symposium on Educational Advances
  in Artificial Intelligence (EAAI-18), New Orleans, Louisiana, USA, February
  2-7, 2018}, pages 3109--3118. {AAAI} Press, 2018.

\bibitem[\protect\citeauthoryear{Han \bgroup \em et al.\egroup
  }{2018}]{han2018co}
Bo~Han, Quanming Yao, Xingrui Yu, Gang Niu, Miao Xu, Weihua Hu, Ivor Tsang, and
  Masashi Sugiyama.
\newblock Co-teaching: Robust training of deep neural networks with extremely
  noisy labels.
\newblock {\em arXiv preprint arXiv:1804.06872}, 2018.

\bibitem[\protect\citeauthoryear{Han \bgroup \em et al.\egroup
  }{2020}]{DBLP:journals/corr/abs-2011-04406}
Bo~Han, Quanming Yao, Tongliang Liu, Gang Niu, Ivor~W Tsang, James~T Kwok, and
  Masashi Sugiyama.
\newblock A survey of label-noise representation learning: Past, present and
  future.
\newblock {\em arXiv preprint arXiv:2011.04406}, 2020.

\bibitem[\protect\citeauthoryear{Huang \bgroup \em et al.\egroup
  }{2020}]{huang2020self}
Lang Huang, Chao Zhang, and Hongyang Zhang.
\newblock Self-adaptive training: beyond empirical risk minimization.
\newblock {\em Advances in Neural Information Processing Systems}, 33, 2020.

\bibitem[\protect\citeauthoryear{Jiang \bgroup \em et al.\egroup
  }{2018}]{jiang2018mentornet}
Lu~Jiang, Zhengyuan Zhou, Thomas Leung, Li-Jia Li, and Li~Fei-Fei.
\newblock Mentornet: Learning data-driven curriculum for very deep neural
  networks on corrupted labels.
\newblock In {\em International Conference on Machine Learning}, pages
  2304--2313. PMLR, 2018.

\bibitem[\protect\citeauthoryear{Kingma and
  Ba}{2015}]{DBLP:journals/corr/KingmaB14}
Diederik~P. Kingma and Jimmy Ba.
\newblock Adam: {A} method for stochastic optimization.
\newblock In Yoshua Bengio and Yann LeCun, editors, {\em 3rd International
  Conference on Learning Representations, {ICLR} 2015, San Diego, CA, USA, May
  7-9, 2015, Conference Track Proceedings}, 2015.

\bibitem[\protect\citeauthoryear{Li \bgroup \em et al.\egroup
  }{2020a}]{li2020dividemix}
Junnan Li, Richard Socher, and Steven~CH Hoi.
\newblock Dividemix: Learning with noisy labels as semi-supervised learning.
\newblock {\em arXiv preprint arXiv:2002.07394}, 2020.

\bibitem[\protect\citeauthoryear{Li \bgroup \em et al.\egroup
  }{2020b}]{DBLP:conf/aistats/LiSO20}
Mingchen Li, Mahdi Soltanolkotabi, and Samet Oymak.
\newblock Gradient descent with early stopping is provably robust to label
  noise for overparameterized neural networks.
\newblock In Silvia Chiappa and Roberto Calandra, editors, {\em The 23rd
  International Conference on Artificial Intelligence and Statistics, {AISTATS}
  2020, 26-28 August 2020, Online [Palermo, Sicily, Italy]}, volume 108 of {\em
  Proceedings of Machine Learning Research}, pages 4313--4324. {PMLR}, 2020.

\bibitem[\protect\citeauthoryear{Liu \bgroup \em et al.\egroup }{2019}]{lupus}
Xiao Liu, Wei Zhang, Ming Zhao, Longfei Liu, Limin Liu, Jinghua Wu, Shuangyan
  Luo, Longlong Wang, Zijun Wang, Liya Lin, Yan Liu, Shiyu Wang, Yang Yang,
  Lihua Luo, Juqing Jiang, Xie Wang, Yixin Tan, Tao Li, Bochen Zhu, and Qianjin
  Lu.
\newblock T cell receptor β repertoires as novel diagnostic markers for
  systemic lupus erythematosus and rheumatoid arthritis.
\newblock {\em Annals of the Rheumatic Diseases}, 78:annrheumdis--2019, 05
  2019.

\bibitem[\protect\citeauthoryear{Lukasik \bgroup \em et al.\egroup
  }{2020}]{lukasik2020does}
Michal Lukasik, Srinadh Bhojanapalli, Aditya Menon, and Sanjiv Kumar.
\newblock Does label smoothing mitigate label noise?
\newblock In {\em International Conference on Machine Learning}, pages
  6448--6458. PMLR, 2020.

\bibitem[\protect\citeauthoryear{Ostmeyer \bgroup \em et al.\egroup
  }{2017}]{stat}
Jared Ostmeyer, Scott Christley, William Rounds, Inimary Toby, Benjamin
  Greenberg, Nancy Monson, and Lindsay Cowell.
\newblock Statistical classifiers for diagnosing disease from immune
  repertoires: A case study using multiple sclerosis.
\newblock {\em BMC Bioinformatics}, 18, 09 2017.

\bibitem[\protect\citeauthoryear{Ostmeyer \bgroup \em et al.\egroup
  }{2020a}]{biophysi}
Jared Ostmeyer, Elena Lucas, Scott Christley, Jayanthi Lea, Nancy Monson,
  Jasmin Tiro, and Lindsay Cowell.
\newblock Biophysicochemical motifs in t cell receptor sequences as a potential
  biomarker for high-grade serous ovarian carcinoma.
\newblock {\em PLOS ONE}, 15:e0229569, 03 2020.

\bibitem[\protect\citeauthoryear{Ostmeyer \bgroup \em et al.\egroup
  }{2020b}]{10.1371/journal.pone.0229569}
Jared Ostmeyer, Elena Lucas, Scott Christley, Jayanthi Lea, Nancy Monson,
  Jasmin Tiro, and Lindsay~G. Cowell.
\newblock Biophysicochemical motifs in t cell receptor sequences as a potential
  biomarker for high-grade serous ovarian carcinoma.
\newblock {\em PLOS ONE}, 15(3):1--17, 03 2020.

\bibitem[\protect\citeauthoryear{Pereyra \bgroup \em et al.\egroup
  }{2017}]{pereyra2017regularizing}
Gabriel Pereyra, George Tucker, Jan Chorowski, {\L}ukasz Kaiser, and Geoffrey
  Hinton.
\newblock Regularizing neural networks by penalizing confident output
  distributions.
\newblock {\em arXiv preprint arXiv:1701.06548}, 2017.

\bibitem[\protect\citeauthoryear{Rolnick \bgroup \em et al.\egroup
  }{2017}]{DBLP:journals/corr/RolnickVBS17}
David Rolnick, Andreas Veit, Serge~J. Belongie, and Nir Shavit.
\newblock Deep learning is robust to massive label noise.
\newblock {\em CoRR}, abs/1705.10694, 2017.

\bibitem[\protect\citeauthoryear{Sidhom \bgroup \em et al.\egroup
  }{2021}]{sidhom2021deeptcr}
John-William Sidhom, H~Benjamin Larman, Drew~M Pardoll, and Alexander~S Baras.
\newblock Deeptcr is a deep learning framework for revealing sequence concepts
  within t-cell repertoires.
\newblock {\em Nature communications}, 12(1):1--12, 2021.

\bibitem[\protect\citeauthoryear{Song \bgroup \em et al.\egroup
  }{2020}]{song2020learning}
Hwanjun Song, Minseok Kim, Dongmin Park, Yooju Shin, and Jae-Gil Lee.
\newblock Learning from noisy labels with deep neural networks: A survey.
\newblock {\em arXiv preprint arXiv:2007.08199}, 2020.

\bibitem[\protect\citeauthoryear{Sá-Nunes}{2021}]{overview}
Anderson Sá-Nunes.
\newblock {\em Overview of Immune Responses}, pages 1--11.
\newblock 10 2021.

\bibitem[\protect\citeauthoryear{Tarvainen and
  Valpola}{2017}]{tarvainen2017mean}
Antti Tarvainen and Harri Valpola.
\newblock Mean teachers are better role models: Weight-averaged consistency
  targets improve semi-supervised deep learning results.
\newblock {\em arXiv preprint arXiv:1703.01780}, 2017.

\bibitem[\protect\citeauthoryear{Thomas \bgroup \em et al.\egroup
  }{2014}]{tracking}
Niclas Thomas, Katharine Best, Mattia Cinelli, Shlomit Reich-Zeliger, Hilah
  Gal, Eric Shifrut, Asaf Madi, Nir Friedman, John Shawe-Taylor, and Benny
  Chain.
\newblock Tracking global changes induced in the cd4 t-cell receptor repertoire
  by immunization with a complex antigen using short stretches of cdr3 protein
  sequence.
\newblock {\em Bioinformatics (Oxford, England)}, 30, 08 2014.

\bibitem[\protect\citeauthoryear{Vaswani \bgroup \em et al.\egroup
  }{2017}]{DBLP:conf/nips/VaswaniSPUJGKP17}
Ashish Vaswani, Noam Shazeer, Niki Parmar, Jakob Uszkoreit, Llion Jones,
  Aidan~N. Gomez, Lukasz Kaiser, and Illia Polosukhin.
\newblock Attention is all you need.
\newblock In Isabelle Guyon, Ulrike von Luxburg, Samy Bengio, Hanna~M. Wallach,
  Rob Fergus, S.~V.~N. Vishwanathan, and Roman Garnett, editors, {\em Advances
  in Neural Information Processing Systems 30: Annual Conference on Neural
  Information Processing Systems 2017, December 4-9, 2017, Long Beach, CA,
  {USA}}, pages 5998--6008, 2017.

\bibitem[\protect\citeauthoryear{Wang \bgroup \em et al.\egroup
  }{2018}]{wang2018revisiting}
Xinggang Wang, Yongluan Yan, Peng Tang, Xiang Bai, and Wenyu Liu.
\newblock Revisiting multiple instance neural networks.
\newblock {\em Pattern Recognition}, 74:15--24, 2018.

\bibitem[\protect\citeauthoryear{Widrich \bgroup \em et al.\egroup
  }{2020}]{DBLP:conf/nips/WidrichSPRGHBSG20}
Michael Widrich, Bernhard Sch{\"a}fl, Milena Pavlovi{\'c}, Hubert Ramsauer,
  Lukas Gruber, Markus Holzleitner, Johannes Brandstetter, Geir~Kjetil Sandve,
  Victor Greiff, Sepp Hochreiter, et~al.
\newblock Modern hopfield networks and attention for immune repertoire
  classification.
\newblock {\em Advances in Neural Information Processing Systems},
  33:18832--18845, 2020.

\bibitem[\protect\citeauthoryear{Wong and
  Li}{2022}]{DBLP:journals/bioinformatics/WongL22}
Christina Wong and Bo~Li.
\newblock Autocat: automated cancer-associated tcrs discovery from tcr-seq
  data.
\newblock {\em Bioinform.}, 38(2):589--591, 2022.

\bibitem[\protect\citeauthoryear{Zhang \bgroup \em et al.\egroup
  }{2021}]{GIANA}
Hongyi Zhang, Xiaowei Zhan, and Bo~Li.
\newblock Giana allows computationally-efficient tcr clustering and
  multi-disease repertoire classification by isometric transformation.
\newblock {\em Nature Communications}, 12:4699, 08 2021.

\bibitem[\protect\citeauthoryear{Zhou \bgroup \em et al.\egroup
  }{2021}]{zhou2021robust}
Tianyi Zhou, Shengjie Wang, and J~Bilmes.
\newblock Robust curriculum learning: From clean label detection to noisy label
  self-correction.
\newblock In {\em Proceedings of the International Conference on Learning
  Representations, Lisbon, Portugal}, pages 28--29, 2021.

\end{thebibliography}

\clearpage

\end{document}